\documentclass{article}
\usepackage{spconf,amsmath,graphicx}
\usepackage{enumitem}
\usepackage{multirow}
\usepackage{graphicx}
\usepackage{float}

\title{Attention-based Memory Selection  Recurrent Network \\ for Language Modeling} 
\name{Da-Rong Liu, Shun-Po Chuang, Hung-yi Lee}
\address{National Taiwan University}
\begin{document}
\ninept
\maketitle
\begin{abstract}

Recurrent neural networks (RNNs) have achieved great success in language modeling.
However, since the RNNs have fixed size of memory, their memory cannot store all the information about the words it have seen before in the sentence, and thus the useful long-term information may be ignored when predicting the next words.
In this paper, we propose Attention-based Memory Selection Recurrent Network (AMSRN), in which the model can review the information stored in the memory at each previous time step and select the relevant information to help generate the outputs.
In AMSRN, the attention mechanism finds the time steps storing the relevant information in the memory, and memory selection determines which dimensions of the memory are involved in computing the attention weights and from which the information is extracted.
In the experiments, AMSRN outperformed long short-term memory (LSTM) based language models on both English and Chinese corpora.
Moreover, we investigate using entropy as a regularizer for attention weights and visualize how the attention mechanism helps language modeling.

\end{abstract}
\begin{keywords}
Language Modeling, Recurrent Network, Attention Model
\end{keywords}
\section{Introduction}
\label{sec:intro}

Recurrent neural networks (RNNs)~\cite{elman1990finding} have been shown to perform well in many sequence modeling tasks\cite{sutskever2014sequence}. In RNNs, the gated memory cells like long short-term memory (LSTM)~\cite{hochreiter1997long} and  gated recurrent unit (GRU)~\cite{chung2014empirical} are widely used.
The attention mechanism has been applied on RNN models. 
Neural Turing Machine (NTM)~\cite{graves2014neural} is one of the examples.
The idea of attention mechanism is to let the model automatically find the related part of information from memory (usually represented as a vector sequence), and use the information to obtain the results. 
Attention mechanism shows promising results on many tasks including machine translation\cite{luong2015effective,cohn2016incorporating,bahdanau2014neural}, caption generation\cite{xu2015show,jin2015aligning} and question answering~\cite{yang2015stacked,chen2015abc,Shih_2016_CVPR,das2016human}.

Language modeling has been recognized as an important task in human language processing. 
The statistical models such as N-gram language model~\cite{rosenfeld2000two,jelinek1991up} were widely used to solve this task.
Recently, RNNs are introduced in language modeling~\cite{mikolov2010recurrent,mikolov2011extensions,gangireddy2014feed,liu2014efficient,zhao2014bilingual,chien2016bayesian,mohamed2015deep,li2015constructing} and have shown great improvement compared to the traditional counterpart.
However, since the RNNs have fixed size of memory, their memory cannot store all the information about the words have seen in the sentence, and thus the useful long-term information may be ignored when predicting the next words.
By making the RNN models have the ability to review the information obtained in every previous time step, the attention mechanism  improves RNN model.
The examples of integrating attention mechanism and LSTM-based RNN model for  language modeling are Long Short-term Memory-Network (LSTMN)~\cite{cheng2016long}  and Recurrent Memory Network (RMN)~\cite{tran2016recurrent}.
LSTMN uses an expandable hidden memory to explicitly store every past memory segments, making use of all the previous values in the memory to compute every update and generate the results. 
RMN uses the hidden memory of LSTM to generate the attention weights, and then uses the attention weights and another trainable memory to generate the outputs. 
Both LSTMN and RMN are shown to outperform original LSTM on language modeling.

In this paper, we propose Attention-based Memory Selection Recurrent Network (AMSRN), a novel RNN architecture that applies the attention mechanism on LSTM. 
In AMSRN, the attention mechanism extracts the relevant information from the LSTM memory states in all the previous time steps for predicting the next word. 
The information in different dimensions of LSTM memory states has different degrees of involvement in attention weight generation and relevant information extraction.
The degree of the involvement for each dimension is different for each time step.
The memory selection mechanism is automatically learned from data.
In this paper, we mainly make the following contributions:
\begin{enumerate}[labelindent=\parindent,leftmargin=*]
\item We investigate different ways of integrating LSTM and attention mechanism. 
The experimental results show that the attention mechanism helps the LSTM language model on three different corpora including English and Chinese.
\item Different from LSTMN which makes some modification on the way LSTM updates the memory, in AMSRN the attention mechanism is stacked on original LSTM, so the architecture of the original LSTM has remained.
Therefore, the LSTM part in AMSRN can be initialized by a typical LSTM language model.
\item 
In RMN, there are two sets of memory, one for computing the attention weights and the other for extracting the information.
On the other hand, the proposed model learns to determine which memory dimensions should be involved more in computing the attention weights and which should be considered more when extracting the information, and the role of each dimension can be different at different time steps. 
From this point of view, RMN can be considered as a special case of the proposed model. 
The experimental results show that the proposed model has more stable performance across different corpora than RMN. 
\item We investigate to use entropy as regularizer for attention weights. 
\item Finally, we make a visualization analysis of how the attention mechanism helps language modeling.
\end{enumerate}

\section{Attention-based Memory Selection Recurrent Network}
\label{sec:pagestyle}

\begin{figure}[tb]
\centering
\includegraphics[width=1.05\linewidth]{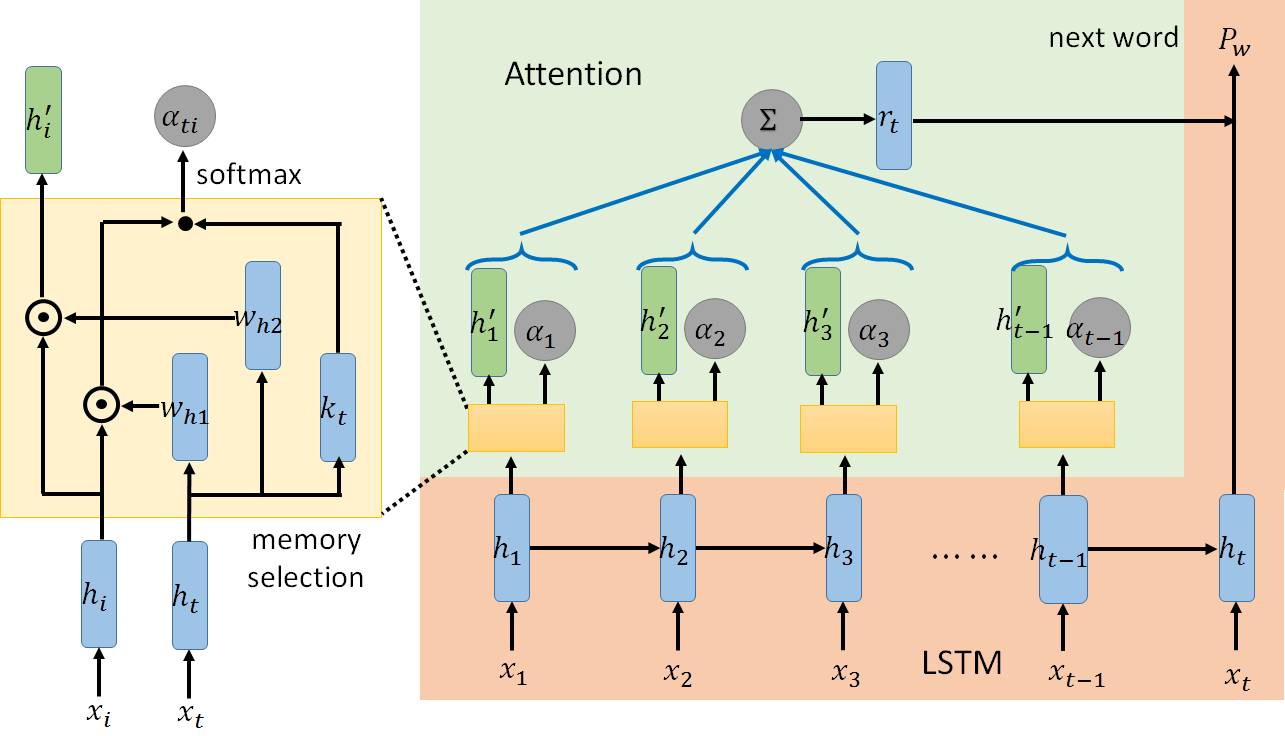}
\caption{The overall structure of the proposed Attention-based Memory Selection Recurrent Network (AMSRN) model.
The notation $\odot$ in the figure represents elementwise multiplication of two vectors.
The notation $\bullet$ represents inner product.}
\label{fig:overview}
\end{figure}

The overall structure of the proposed Attention-based Memory Selection Recurrent Network (AMSRN) is shown in Fig.~\ref{fig:overview}.
AMSRN consists of two major parts: the typical LSTM described in Section~\ref{subsec:lstm} and the attention mechanism module stacking on LSTM in Section~\ref{subsec:att}.
The LSTM reads through the input word sequence, and stores the hidden layer outputs generated at each time step.
The attention mechanism module takes the stored information as input, and generates a vector relevant to the prediction of the next words.
Then the relevant vector and the current   LSTM hidden state  are used to generate the distributions for the next words.
In the attention mechanism module, the memory selection is applied on the LSTM memory to determine which dimensions of LSTM hidden states should be involved in computing the attention weights and extracting relevant information, which will be described in Section~\ref{subsec:selection}.
Finally, the attention weights can be regularized by their entropy in Section~\ref{subsec:reg}.

\subsection{LSTM}  \label{subsec:lstm}
The input of the LSTM is a sequence of words represented by 1-of-N encoding, $\{x_1,x_2, \cdots x_t, \cdots \}$, at the bottom of Fig.~\ref{fig:overview}.
At each time step $t$, the hidden layer output of the LSTM is a $d$-dimensional  vector $h_t$, where $d$ is the number of memory cells in LSTM, and $h_t$ would be stored for further use.
Therefore, at the time step $t$ (when the model has read the first $t$ words in the sentence in question), the information stored is $M_{t}$,
\begin{equation}
M_{t}= [ h_{0}, h_{1}, \cdots h_{t-1} ], \label{eq:memory}
\end{equation}
where $h_0$ is the initial state of the LSTM.
$M_{t}$ in (\ref{eq:memory}) is a $d \times t$ matrix, which grows as  $t$ increases. 
The attention module will extract the information from $M_{t}$.


\subsection{Attention Mechanism} \label{subsec:att}

In the attention mechanism, the memory selection module generates two $d$-dimensional vectors, $w_{h1}$ and $w_{h2}$, from the current LSTM state $h_t$. 
$w_{h1}$ and $w_{h2}$ are used to select the stored information.
Here all the elements in $w_{h1}$  and $w_{h2}$ are between $0$ and $1$.
How to generate $w_{h1}$  and $w_{h2}$ will be described in the next subsection.


Then the current hidden state $h_{t}$ and the two memory selection vectors, $w_{h1}$  and $w_{h2}$,  are used to extract the relevant information, represented as a $d$-dimensional vector $r_t$, from   $M_t$ in (\ref{eq:memory}).
The model first generates a $d$-dimensional vector $k_{t}$ from the current hidden state $h_t$ as the `key'  for attention weight generation, 
\begin{equation}
k_{t} = W_{kh} h_{t} + b_{k},
\end{equation}
where the $d \times d$ matrix $W_{kh}$ and $d$-dimensional vector $b_{k}$ are network parameters to be learned.
Then the inner product similarity $e_{ti}$ between the key $k_t$ and each $h_i$ in $M_t=[h_0, h_1, h_2, \cdots h_i, \cdots h_{t-1}]$ is computed. 
\begin{equation}
e_{ti} =  (h_{i} \odot w_{h1}) \bullet k_{t}, \label{eq:dot}
\end{equation}
where $\odot$ denotes the elementwise multiplication, and $\bullet$ denotes the inner product.
By multiplying each element in $h_{i}$ by the corresponding element in $w_{h1}$ (that is, $h_{i} \odot w_{h1}$ in (\ref{eq:dot})), the model determines the degree of each  dimension of $h_i$ involved in computing the similarity (for example, the dimension multiplied by $0$ would be totally ignored in generating the attention weights).
The similarity $e_{ti}$ is further normalized by softmax normalization to obtain the attention weights $\alpha_{ti}$,
\begin{equation}
\alpha_{ti} = \frac{exp(e_{ti})}{\sum_{i=0}^{t-1}exp(e_{ti})}.
\end{equation}
To generate the relevant vector $r_{t}$, each $h_{i}$ is selected by $w_{h2}$ to obtain $h_{i}^{\prime}$,
\begin{equation}
h_{i}^{\prime} = h_{i} \odot w_{h2},
\end{equation}
in which the degree each dimension of $h_i$ is involved in extracting the relevant vector $r_{t}$ is determined.
Finally,  $r_{t}$ is generated by the weighted sum of  $h_{i}^{\prime}$ according to $\alpha_{ti}$,
\begin{equation}
r_{t}= \sum_{i=0}^{t-1} \alpha_{ti}h_{i}^{\prime}. \label{eq:extract} 
\end{equation}
The attention vector $r_{t}$ and the hidden state $h_{t}$ predicts the distribution of the next word $P_{w}$,
\[
P_{w}=softmax(W_{ph}h_{t} + W_{pr}r_{t} + b_{p}),
\]
where $W_{ph}$, $W_{pr}$ and $b_{p}$ are network parameters to be learned, and $softmax$ is the softmax activation function.
The cost $C$ to be minimized by optimizing the network parameters is the cross-entropy between the word distribution $P_{w}$ and the reference distribution for all the words in the training set.

\subsection{Memory Selection} \label{subsec:selection}
In this paper, we investigate three different ways to obtain  $w_{h1}$ used in (\ref{eq:dot}) and $w_{h2}$ in (\ref{eq:extract}):
\begin{enumerate}[labelindent=\parindent,leftmargin=*]
\item  $w_{h1}$  and $w_{h2}$ are generated independently. The current state of LSTM, $h_{t}$, is passed into two different fully connected layers with sigmoid activation function to generate $w_{h1}$ and $w_{h2}$ as below,
\begin{align*}
w_{h1}=sigmoid(W_{hh1}h_{t} + b_{h1})\\
w_{h2}=sigmoid(W_{hh2}h_{t} + b_{h2}), \label{eq:generate_key}
\end{align*}
where the $W_{hh1}$, $b_{h1}$, $W_{hh2}$ and $b_{h2}$ denote the weights and the biases of the fully connected layer.
\item The two vectors $w_{h1}$ and $w_{h2}$ are forced to be the same.
$w_{h1}$ is generated by the same way as the first approach, and the model simply sets $w_{h2} = w_{h1}$.
\item The only difference between the third and the second approaches is that here we set $w_{h1}=\mathbf{1}-w_{h2}$, where $\mathbf{1}$ is a $d$-dimensional vector with all ones, and `$-$' here represents elementwise subtraction.
The inspiration of the third approach is that in the attention models the memory used to generate attention distribution and the memory used to generate the final attention vector can be different~\cite{tran2016recurrent,sukhbaatar2015end}. 
Therefore, by constraining the sum of the two weights, $w_{h1}$ and $w_{h2}$, it simulates the situation that there are two different sets of memory for attention weights and information extraction respectively.
\end{enumerate}

\subsection{Regularizer}   \label{subsec:reg}
When training model, a regularization term is usually used to prevent overfitting. 
For example, the two-norm of the model parameters are widely used as a regularizer.
Here we investigate to use the entropy of the attention weights as the regularization term~\cite{grandvalet2004semi}. 
The purpose of using entropy as regularizer is because	only part of the information in the previous steps is relevant to the prediction of the next word.
Therefore, the attention weights that extract useful information from the previous time steps are sparse.
The entropy regularizer to keep the attention weights sparse is designed as below.
\begin{equation}
L_{reg} = \sum_{u} \sum_{t=1}^{T_u} \sum_{i=0}^{t-1} {-w_{ti}\log{w_{ti}}}, \label{eq:reg}
\end{equation}
where $u$ is a sentence in the training corpus, and $T_u$ is the length of $u$.
$w_{ti}$ in (\ref{eq:reg}) denotes the attention weight of $h_i$ at the time step $t$ when reading sentence $u$, and $\sum_{i=0}^{t-1} {-w_{ti}\log{w_{ti}}}$ is the entropy of the attention weights obtained at the time step $t$.
With the regularization term, the network is learned to minimize $C + \lambda L_{reg} $, where $C$ has been mentioned in Subsection~\ref{subsec:att} and $\lambda$ is determined by a validation set. 

\begin{figure*}[tb]
\centering
\includegraphics[width=0.75\linewidth]{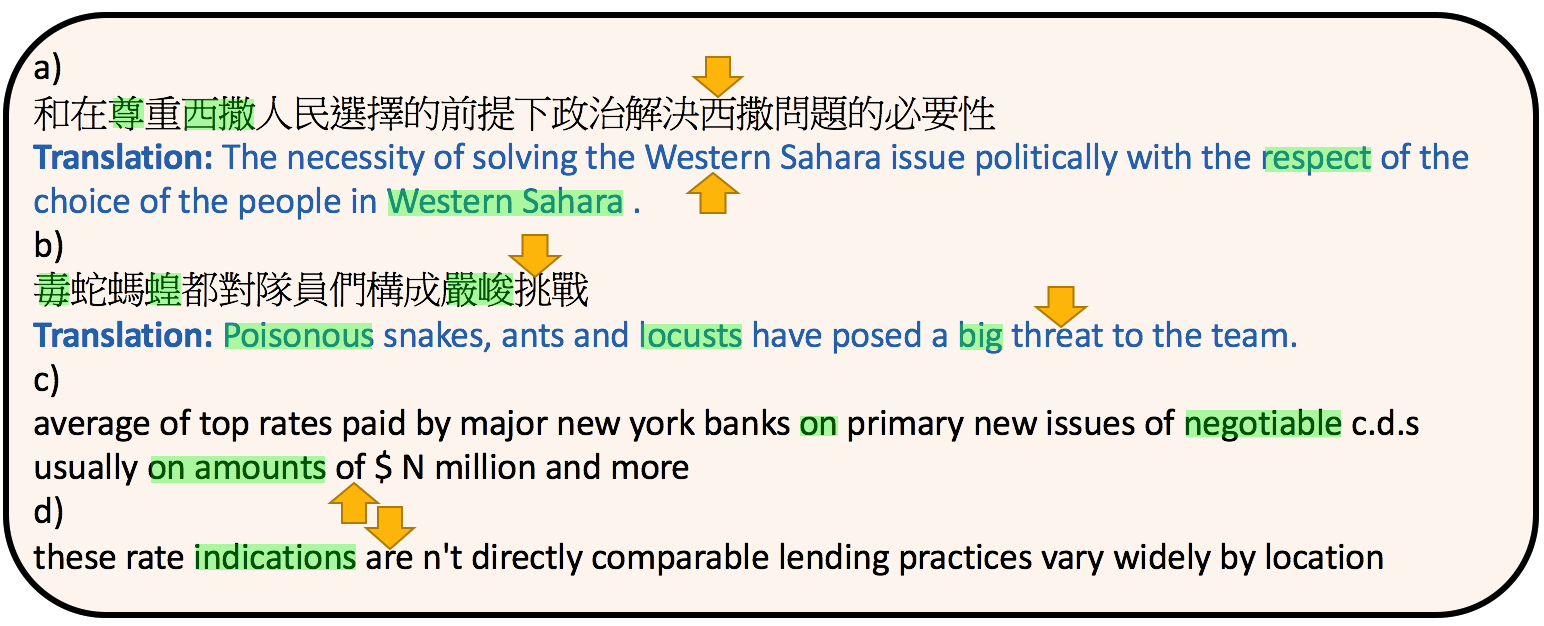}
\caption{The visualization of attention weights. 
The arrows point to the words to be predicted. 
The words with attention are highlighted.
For the Chinese examples, \textbf{the order of the Chinese characters in a sentence will not be the same as its English translation}. 
There are four examples in the figure:
(a) Trigger: attend to the same place name shown before in the sentence.
(b) Causal Relationship: There are poisonous insects, so the task is a challenge. 
(c) Phrases:'on amount of' is a phrase, so the first two words will help determine the last word.
(d) Grammar: we should use 'are' right after a plural noun. }
\label{fig:vis}
\end{figure*}

\section{Experiments}

\subsection{Experimental Setups} 
We tested the proposed model on two English data sets and one Chinese data set. 
The first data set we used is the Penn Treebank Corpus~\cite{taylor2003penn}, which is a widely used data set to evaluate the effectiveness of a language model. 
It contains about 40K training sentences, 3K validation sentences and 4K testing sentences. 
The other English data set we used is from the Switchboard Corpus\cite{godfrey1992switchboard}.
 Switchboard is a Telephone Speech Corpus which collect two-sided telephone conversations among speakers in the United States. 
 We used about 945K sentences for training, 10K for validation and about 5.2K for testing.  
 For Chinese, we used Chinese Gigaword data set\cite{graff2005chinese} to evaluate the model. 
 Chinese Gigaword data set consists of around 25K Chinese news articles. 
After parsing, there are 531K sentences for training, 165K for validation and about 260K for testing. 
 Table 1 summaries the statistics of the three data sets we used in the following experiments.
 The perplexities (PPLs) on the testing data sets are used to evaluate different methods.
\begin{table}[]
	\centering
	\caption{The statistics of the three data sets we used in the following experiments.}
	\label{my-label}
	\begin{tabular}{|c|c|c|c|c|c|c|}
		\hline
Corpus	&	Lang & \multicolumn{1}{r|}{train} & \multicolumn{1}{r|}{dev} & test & $\left| s \right|$ & $\left| v \right|$ \\ \hline
		\bf PT &Eng       & \multicolumn{1}{r|}{40K}   & \multicolumn{1}{r|}{3K}  & 4K  &  21.1   & 9999    \\ \hline
		\bf SB &Eng        & \multicolumn{1}{r|}{945K}      & \multicolumn{1}{r|}{10K}    &  5.2K    & 10.39   &  47283 \\ \hline
		\bf{GW}     & Chi      & \multicolumn{1}{r|}{531K}      & \multicolumn{1}{r|}{165K}    &  260K    & 10.79   &  5123 \\ \hline
		\multicolumn{7}{l}{\footnotesize \textit{$\left| s \right|$ } denotes the average number of words in the sentences.}\\
		\multicolumn{7}{l}{\footnotesize \textit{$\left| v \right|$ } denotes the size of the vocabulary.}\\
		\multicolumn{7}{l}{\footnotesize   PT denotes Penn Treebank Corpus.} \\
		\multicolumn{7}{l}{\footnotesize  SB denotes Switchboard Corpus.}\\
		\multicolumn{7}{l}{\footnotesize  GW denotes Gigaword Corpus.}\\
	\end{tabular}
\end{table}

Although it is possible to train AMSRN model from scratch, since the AMSRN model contains a LSTM part, it is possible to initialize the LSTM part by a LSTM langauge model.
Therefore, in the following experiments, the AMSRN is always pretrained by a LSTM language model.
Because of the limited computing resource, we fixed both the dimension of the LSTM hidden state and the embedding layer to be 50.

\subsection{Memory Selection Methods}
In this experiment, we investigate different memory selection methods for generating $w_{h1}$ and $w_{h2}$ respectively. 
The three methods in Subsection~\ref{subsec:selection} are (1) generating $w_{h1}$ and $w_{h2}$ are independently, (2) setting $w_{h1}=w_{h2}$, and (3) setting $w_{h1}=\mathbf{1}-w_{h2}$.
The results of the three methods are shown in Table 2.
Generating $w_{h1}$ and $w_{h2}$ independently leads to the worst result (133.80).
This may be because this approach need twice parameters compared with the other two.
Constraining $w_{h1}$ and $w_{h2}$ to be the same is better than making them complementary (133.36 v.s. 133.62).
The results suggest that probably the information for computing similarity and information extraction are contained in the same dimension of the LSTM hidden states.
Since the second method achieves the best result, it is used in the following experiments.



\begin{table}[]
\centering
\caption{Comparison of the three memory selection approaches in Subsection~\ref{subsec:selection} on the Penn Treebank corpus.}
\label{my-label}
\begin{tabular}{|r|c|l|}
\hline
\textbf{weight relation} & \multicolumn{2}{r|}{perplexity} \\ \hline
(1) $w_{h1}$ and $w_{h2}$ are independent   & \multicolumn{2}{c|}{133.80}     \\ \hline
(2) $w_{h1}=w_{h2}$     & \multicolumn{2}{c|}{133.36}     \\ \hline
(3) $w_{h1}=\mathbf{1}-w_{h2}$    & \multicolumn{2}{c|}{133.62}     \\ \hline
\end{tabular}
\end{table}
 
\subsection{Comparison of Different Models}
The experimental results of different models are shown in Table 3.
Columns (1), (2) and (3) are the results on Penn Treebank Corpus, Switchboard Corpus and Chinese Gigaword data set, respectively.
Experiments were done step by step. 
First, a typical LSTM language model was trained, and PPLs of the LSTM model on the testing sets are in row (A).
Then in row (B) the attention module was added on top of LSTM but without memory selection (or all the elements in $w_{h1}$ and $w_{h2}$ are one) and entropy regularizer.
It is found that attention mechanism was helpful on both Penn Treebank and Switchboard (Rows (B) v.s. (A) on Columns (1) and (2)), but it does not improve the LSTM on Chinese Gigaword (Rows (B) v.s. (A) on columns (3)).
In row (C), we show the results of wit memory selection based on the second approach in Subsection~\ref{subsec:selection}. 
We found that memory selection is essential for attention mechanism here. 
With memory selection, attention-based model outperformed LSTM on all the three corpora (Rows (C) v.s. (B)).
Then the entropy regularization for the attention weights was applied on the attention-based model with memory selection.
The results are in row (D).
The results of entropy regularization are mixed. 
It improved the performance on Penn Treebank, but degrades the performance on the rest two corpora (Rows (D) v.s. (C)).
The experimental results suggest that the assumption of sparse attention weights is probably not very accurate.

We further compare the proposed model with another two attention-based language model, recurrent memory network (RMN)~\cite{tran2016recurrent} and Recurrent-Memory-Recurrent (RMR)~\cite{tran2016recurrent}.
Comparing LSTM with the two attention-based model in the literature, the conclusion is also mixed.
RMN and RMR outperformed drastically the two English corpora (Rows (E), (F) v.s. (A) on Columns (1) and (2)), but contrary conclusion is obtained on the Chinese corpus (Rows (E), (F) v.s. (A) on Column (3)).
This is probably RMN and RMR have only be verified on English, German, and Italian, and there are some special techniques on Chinese that should be specially considered.
The proposed approach consistently improves LSTM, and better than RMN and RMR on Chinese, but worse than them on English corpora.
The proposed model seems to be more robust across different corpora, but the improvements are limited.


\begin{table}[]
	\centering
	\caption{Perplexity Evaluation.}
	\label{my-label}
	\begin{tabular}{|r|r|r|r|r|}
		\hline
		\textbf{model}       & \multicolumn{1}{r|}{(1) PT}    & \multicolumn{1}{r|}{(2) SB} & \multicolumn{1}{r|}{(3) GW}  \\ \hline
		(A) LSTM                 & \multicolumn{1}{r|}{143.31} &\multicolumn{1}{r|}{93.90}   &\multicolumn{1}{r|}{94.03}   \\ \hline
		(B) LSTM+att             & \multicolumn{1}{r|}{134.09} & \multicolumn{1}{r|}{93.74}   &\multicolumn{1}{r|}{102.04}   \\ \hline
		(C) LSTM+att+select         & \multicolumn{1}{r|}{133.36} & \multicolumn{1}{r|}{92.49}  &\multicolumn{1}{r|}{86.85}    \\ \hline
		(D) LSTM+att+select+entropy & \multicolumn{1}{r|}{131.43} & \multicolumn{1}{r|}{99.79}     &\multicolumn{1}{r|}{88.25} \\ \hline
		(E) RMN~\cite{tran2016recurrent} & \multicolumn{1}{r|}{123.32} & \multicolumn{1}{r|}{64.41}     &\multicolumn{1}{r|}{121.28} \\ \hline
		(F) RMR~\cite{tran2016recurrent} & \multicolumn{1}{r|}{134.30} & \multicolumn{1}{r|}{71.04}     &\multicolumn{1}{r|}{145.24} \\ \hline
		\multicolumn{4}{l}{\footnotesize att: attention mechanism }\\
		\multicolumn{4}{l}{\footnotesize select: memory selection mechanism }\\
		\multicolumn{4}{l}{\footnotesize entropy: entropy regularizer}\\
	\end{tabular}
\end{table}

\subsection{Analysis}
To illustrate how attention mechanism works, we visualize the attention weights in some sentences.
We first compute the perplexities of each sentence in Gigaword (Chinese) and  Penn Treebank (English) data sets, then select the sentences which improved the most by  the proposed model (Row (C) in Table 3) compared with the LSTM baselines. 
We chose ten sentences from Gigaword (Chinese) and  Penn Treebank data sets, and visualize and analysis the attention weights. 
Four examples are shown in Fig.~\ref{fig:vis}.
In Fig.~\ref{fig:vis}, the arrows point to the words to be predicted, and we highlight the words whose attention weights are higher than a threshold when predicting the words with arrows.
We found that a word will have large attention under one of the following four conditions:
\begin{enumerate}[labelindent=\parindent,leftmargin=*]
	\item Trigger (example (a) in Fig.~\ref{fig:vis}: 
	When the information is repeated, the model attends to the part where the same information is mentioned before.
	\item Causal Relationship (example (b)): 
	If $A$ is the cause of $B$, when prediction the words related to $B$, the model will attend to the words related to $A$.
	\item Phrases (example (c)): 
	When predicting a word in the later part of a phrase, the model will attend on the former part of the same phrase.
	\item Grammar (example (d)): 
	Some grammar rules are considered by the attention-based model.
	For example, to predict the word 'are', the model attends on a plural noun. 
\end{enumerate}

\section{Concluding Remarks}
In this paper, we propose Attention-based Memory Selection Recurrent Network (AMSRN) for language modeling and investigate the integration of attention mechanism and LSTM.
The results were verified on two English corpora and a Chinese corpus.
The results show that AMSRN consistently outperformed LSTM-based language model, and memory selection is essential for attention mechanism.
We further visualize how the attention mechanism works in language modeling.
Some questions unresolved in this paper will be studied in the future, for example, the influence of the language characteristics to the attention-based model.



\end{document}